\documentclass{article}
\usepackage{ijcai23}

\usepackage{times}
\usepackage{soul}
\usepackage{url}
\usepackage[hidelinks]{hyperref}
\usepackage[utf8]{inputenc}
\usepackage[small]{caption}
\usepackage{graphicx}
\usepackage{amsmath}
\usepackage{amsthm}
\usepackage{booktabs}
\usepackage{algorithm}
\usepackage{algorithmic}
\usepackage[switch]{lineno}
\usepackage{cuted}
\usepackage{xcolor}
\usepackage{colortbl} 
\usepackage{multirow}

\usepackage{pifont}

\def\figvspace{{\vspace{-3mm}}}

\title{Multimodal Frame-Scoring Transformer for Video Summarization}

\author{
Jeiyoon Park$^{1,4}$
\and
Kiho Kwoun$^2$\and
Chanhee Lee$^{3}$\And
Heuiseok Lim$^1$
\affiliations
$^1$Department of Computer Science and Engineering, Korea University\\
$^2$School of Electrical and Computer Engineering, University of Seoul\\
$^3$Naver Corporation\\
$^4$LLSOLLU
\emails
\{k4ke, limhseok\}@korea.ac.kr,
dev.chocochip@uos.ac.kr,
chanhee.lee@navercorp.com
}

\begin{document}

\maketitle

\begin{abstract}
    As the number of video content has mushroomed in recent years, automatic video summarization has come useful when we want to just peek at the content of the video. However, there are two underlying limitations in generic video summarization task. First, most previous approaches read in just visual features as input, leaving other modality features behind. Second, existing datasets for generic video summarization are relatively insufficient to train a caption generator used for extracting text information from a video and to train the multimodal feature extractors. To address these two problems, this paper proposes the Multimodal Frame-Scoring Transformer (MFST), a framework exploiting visual, text, and audio features and scoring a video with respect to frames. Our MFST framework first extracts each modality features (audio-visual-text) using learned encoders. Then, MFST trains the multimodal frame-scoring transformer that uses multimodal representation based on extracted features as inputs and predicts frame-level scores. Our extensive experiments with previous models and ablation studies on TVSum and SumMe datasets demonstrate the effectiveness and superiority of our proposed method by a large margin in both F1 score and Rank-based evaluation.
\end{abstract}

\section{Introduction}

\textbf{Our Intuition about Video Summarization.} When humans watch a video on YouTube or Netflix, they perceive visual, linguistic, and audio information through various sense organs and know which parts of the video are interesting. To consider whether a scene in a movie is absorbing, for example, we observe characters' facial expressions and actions, recognize background and situation with language, and listen to the characters' utterances and sound effects. Intuitively, humans have access to well-defined scoring function in their mind, using versatile sensory systems, while video summarization models from previous studies did not. 

\begin{figure}[t]
	\centering
	\begin{minipage}[h]{\linewidth}
		\centering
		\includegraphics[width=1\linewidth]{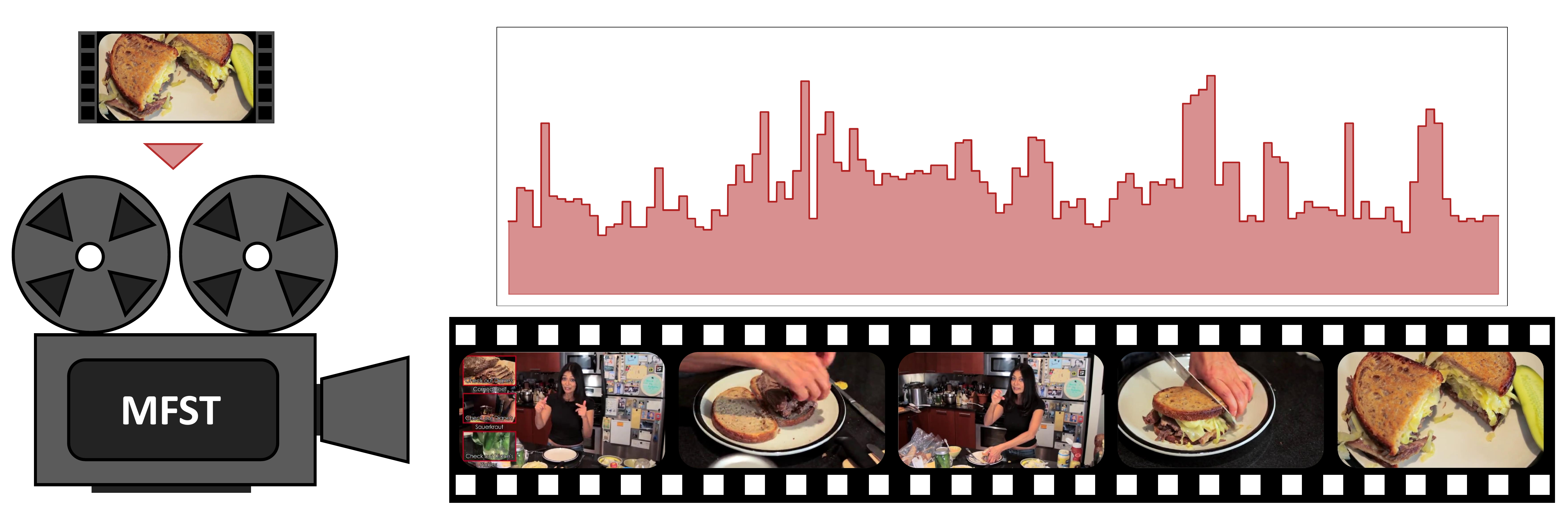}
	\end{minipage}
	\figvspace
	\caption{The proposed MFST takes a video as input and generates frame-level importance scores for video summarization. The \textcolor[RGB]{178,34,34}{red} areas indicate predicted importance score.
	}
	\vspace{-3mm}
	\label{fig:1}
\end{figure}

\textbf{Two Underlying Research challenges.} Video summarization aims to capture key frames using predicted frame-wise importance scores, given datasets as shown in Figure~\ref{fig:1}. Despite its importance and convenience, video summarization has two inherent challenges: (i) Most previous approaches exploit just visual features, leaving other modality features behind \cite{Zhao2018HSARNNHS,10.1007/978-3-030-01258-8_22,10.1007/978-3-030-01237-3_24,10.5555/3504035.3504964,DBLP:conf/aaai/JungCKWK19,8954281,Park2020SumGraphVS,10.1007/978-3-030-58595-2_11,ghauri2021MSVA}. (ii) Since the existing datasets \cite{10.1007/978-3-319-10584-0_33,7299154}, which consist only of videos and frame-level ground truths, relatively insufficient to train a caption generator used for extracting text information from the datasets and to train audio-visual-text feature extractors. Note that frame-wise human-scored video dataset is expensive to obtain compared to other common video datasets. 

Language-attended methods exploit extracted text information from videos and predict importance score based on visual and text features. \cite{BMVC2017_118} jointly combines video summarization and video caption model and trains the recurrent network in end-to-end manner. \cite{NEURIPS2021_7503cfac} leverages language-guided video summarization model given videos and corresponding user query or automatically generated video captions. \cite{https://doi.org/10.48550/arxiv.2201.02494} collects video titles and descriptions for pre-training the language-attended self-supervised learning model. 

Though language-guided approaches alleviate modality issue somewhat, there are underlying constraints in conveying vivid audio features into a video summarization model (e.g., we know there are limits to expressing a beautiful song just with the text \textit{“beautiful song”}). 

\textbf{Our Solutions.} This paper proposes Multimodal Frame-Scoring Transfomer (MFST) to handle the modality combinations and frame-level scoring for video summarization. Note that existing datasets for generic video summarization \cite{10.1007/978-3-319-10584-0_33,7299154} are relatively insufficient to pretrain a dense caption generator and audio, visual, and text feature extractors. We investigate a new multimodal setting where it can mitigate the lack of human-scored videos used for training video summarization model.

To this end, our framework consists of three stages: (i) Generating dense video captions using learned caption generator and extracting each modality features (audio-visual-text) using feature encoders. (ii) Multimodal representation using caption-guided attention mechanism and coarse-grained projection based on modality fusion (iii) Frame-scoring transformer that takes audio-visual-text multimodal representations as input and predicts frame-level scores.  

Our extensive experiments with previous models and ablation studies on TVSum and SumMe datasets demonstrate the effectiveness and superiority of our proposed method by a large margin in both F1 score and Rank-based evaluation.

\textbf{Our Contributions.} The main contributions of this work are summarized as below:

\begin{itemize}
    \item To the best of our knowledge, Our MFST is the first to introduce frame-scoring transformer exploiting multimodal features (audio-visual-text) for generic video summarization task.  
    \item We investigate a new multimodal setting where it can mitigate the lack of human-scored videos used for training generic video summarization model exploiting pretrained modules.
    \item Our empirical study on generic video summarization datasets (TVSum and SumMe) demonstrates that MFST can surpass its all counterparts by nontrivial margins, and attests the effectiveness and superiority of our approach.
\end{itemize}

\section{Related Works}

\textbf{Two broad categories of Video Summarization.} In video summarization task, there are two broad categories of methods: (i) generic video summarization \cite{Park2020SumGraphVS,10.1007/978-3-030-58595-2_11,ghauri2021MSVA,NEURIPS2021_7503cfac,https://doi.org/10.48550/arxiv.2201.02494} and (ii) query-guided video summarization \cite{NEURIPS2021_7503cfac,Wu_2022_CVPR,liu2022umt,jiang2022joint}.

\textbf{Generic Video Summarization.} The First category of methods aim to extract representative frames from original videos using well-defined frame-wise scoring function. Existing models focus on both supervised learning and unsupervised learning. \cite{10.5555/3504035.3504964} designed a reward function which determines diversity and representativeness of generated summaries based on end-to-end reinforcement learning framework. \cite{10.1007/978-3-030-01258-8_22} tried to solve video summarization as a sequence labeling problem based on fully convolutional sequence models. \cite{10.1007/978-3-030-01237-3_24} proposed the retrospective encoder to embed both predicted summary and original video. \cite{Zhao2018HSARNNHS} integrated shot-level segmentation and video summary into a hierarchical RNN. \cite{DBLP:conf/aaai/JungCKWK19} proposed variance loss with variational autoencoder and generative adversarial networks. \cite{8954281} learned a mapping function between raw videos and summarized videos because this kind of dataset is much easier to gain. \cite{Yuan2019CycleSUMCA} proposed a cycle-consistent adversarial networks which consist of frame selector and evaluator. \cite{10.1007/978-3-030-58595-2_11} exploited global and local input decomposition to capture the interdependencies of video frames. To represent a relation graph, \cite{Park2020SumGraphVS} leveraged recursive graph modeling networks. 

Noted that, these methods just employs visual features, leaving other modality features behind. In this paper, we investigate a new multimodal setting for video summarization including audio-visual-text features.  

\textbf{Query-Guided Video Summarization.} The second category of methods find relevant moments according to user-defined query. Unlike generic video summarization, most query-guided models take Query-Focused Video Summarization \cite{Sharghi2016QueryFocusedEV} dataset, UT Egocentric \cite{6247820} dataset, and QVHighlights \cite{lei2021detecting} dataset as input. \cite{Sharghi2017QueryFocusedVS} introduced a parametrized memory network into query-focuesd video summarization. \cite{10.5555/3504035.3504062} proposed a semantic attended network which consists of frame selector and video descriptor. \cite{Kanehira2018ViewpointAwareVS} investigated how to divide videos into groups under the assumption that video summaries extracted from similar videos should be similar. \cite{NEURIPS2021_7503cfac} leverages a framework for handling both generic model and query-guided model given videos and corresponding user query or automatically generated video captions. To effectively address generic queries from different modalities \cite{Wu_2022_CVPR} introduced a graph convolutional networks which is used for both summary module and intent module. \cite{liu2022umt} proposed the unified multimodal transformer to cover different input modality combinations. \cite{jiang2022joint} jointly leveraged video summarization model and moment localization model.

Note that though query-driven approach is necessary because defining salient scenes is often subjective, it is difficult to apply if we do not know the contents of the video or if we do not need a subjective summary (e.g., YouTube video previews). In this paper, we aim to tackle the first category based on a novel multimodal frame-scoring framework.

\textbf{Frame-Scoring Transformer.} Some recent works apply transformer \cite{NIPS2017_3f5ee243} for generating video summaries. \cite{NEURIPS2021_7503cfac} employed transformer for predicting importance score using language-attended representation. \cite{liu2022umt} leveraged modality encoder, query generator, and query decoder using transformer framework. Compared to \cite{NEURIPS2021_7503cfac}, we propose frame-scoring transformer exploiting audio-visual-text modality represenatation.

\section{Our Approach}

\begin{figure*}[t]
	\centering
	\begin{minipage}[h]{\linewidth}
		\centering
		\includegraphics[width=1\linewidth]{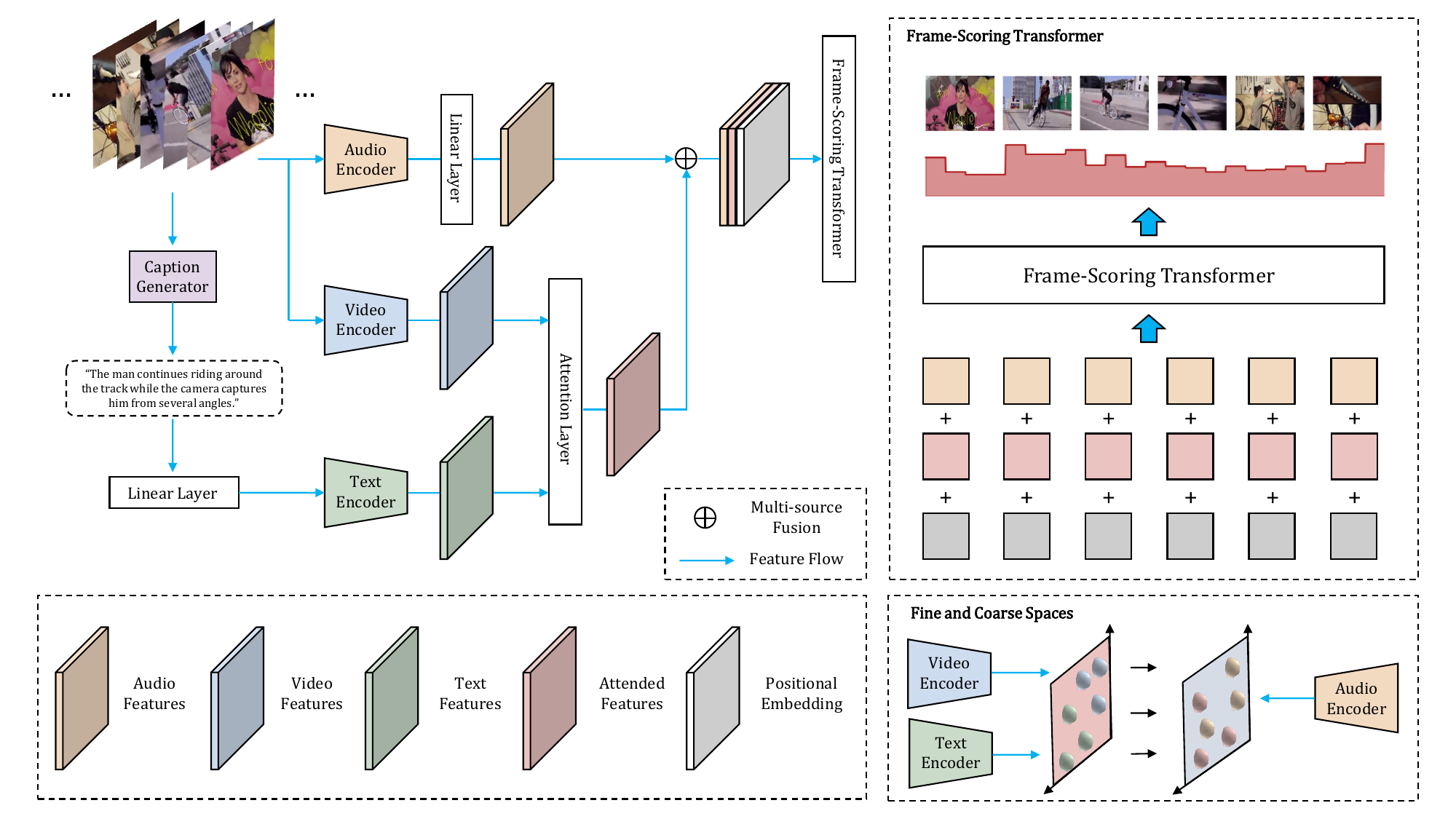}
	\end{minipage}
	\figvspace
	\caption{{Schematic depiction of the proposed Multimodal Frame-Scoring Transformer (MFST) for video summarization.} Given a video, we first generate dense video captions using learned caption generator. To mitigate the lack of human-scored videos, we extract each modality features (audio-visual-text) from videos and generated captions, exploiting learned feature extractors. Then, we calculate text-attended visual representation using attention layer and compute Multi-source fusion across the text-attended features and audio features. Finally we feed the fused representation to frame-scoring transformer with positional encodings at the bottom of transformer encoder and decoder stacks.
	We modify the transformer so that it can use modality-fused representations as input and output predicted frame-level scores.
	}
	\vspace{-3mm}
	\label{fig:2}
\end{figure*}

\textbf{Our Goal.} In this section, we present the proposed MFST framework (as shown in Figure~\ref{fig:2}) with extracted modality features. we consider the standard generic video summarization setting, where given a set of videos $V$ and ground truth frame scores $S_{gt}$, the goal is to minimize the loss $\mathcal{L}_{\theta}$ with respect to predicted frame scores $S$:
\begin{align}
	\hat{\theta} = \arg \min_{\theta}(\mathcal{L}_{\theta}(S_{gt}, S))
\end{align}

\textbf{Multimodal Feature Extraction.} Note that existing datasets for generic video summarization \cite{10.1007/978-3-319-10584-0_33,7299154} are relatively insufficient to train a caption generator and audio, visual, and text feature extractors. Considering that large-scale videos scored by humans are not available, we co-opt a pretrained caption generator and feature extractors for each modality:
\begin{align}
    C = g_{c}(V)\\
    \mathcal{T} = f_{t}(C)\\
    \mathcal{V} = f_{v}(V)\\
    \mathcal{A} = f_{a}(V)
\end{align}
, where $C$ denotes a set of dense video captions, $g_{c}$ is a dense video caption generator \cite{BMT_Iashin_2020}, $f_{t}$ is a CLIP-based text feature extractor \cite{xu-etal-2021-videoclip}, $f_{v}$ is a CLIP-based visual feature extractor \cite{xu-etal-2021-videoclip}, and $f_{a}$ is a learned audio feature extractor in \cite{NEURIPS2020_92d1e1eb}. Here, we use audio-visual-text features at the same time like human to make it easier for our model in understanding video context and demoting peripheral parts.   

\textbf{Fine and Coarse Spaces.} Since our model handles three modality features, $\mathcal{T}, \mathcal{V}$, and $\mathcal{A}$, configuration of modality space should be considered. Fine and coarse spaces (FAC) \cite{10.5555/3495724.3495727} combines the modalities into a common embedding, while preserving fine-grained information and integrating modality features. Inspired by \cite{10.5555/3495724.3495727}, but unlike the study, we first propose to learn visual-text granularities in the FAC \textit{modality embedding graphs}.

Though video caption methods based on automatic speech recognition are useful \cite{hessel-etal-2019-case,10.5555/3495724.3495727,MDVC_Iashin_2020}, most videos in generic video summarization rarely have human dialogues. Note that in real-world applications, not all videos contain human voice, nor we always have to figure out exactly what people are saying for highlight detection (e.g., \textit{croud cheers at a soccer game}). 

In this paper, we co-opt the feature-based video cation generator $g_c$ \cite{BMT_Iashin_2020} to extract dense video captions $C = \{C_{1}, C_{2}, ..., C_{N}\}$. Then, we leverage a fine-grained embedding space where visual features $\mathcal{V} = \{\mathcal{V}_{1}, \mathcal{V}_{2}, ..., \mathcal{V}_{N}\}$ and text features $\mathcal{T} = \{\mathcal{T}_{1}, \mathcal{T}_{2}, ..., \mathcal{T}_{N}\}$ lie. Note that \cite{ModalityGap2022} demonstrates "multimodal video-text pretraining" paradigm can not solve the \textit{modality gap} phenomenon completely which causes performance degradation. In this work, we compute text-attended visual representation using attention layer:  
\begin{align}
	h_{\mathcal{V}\mathcal{T}} = 
    \text{Attention}(h_{\mathcal{V}}, h_{\mathcal{T}})
\end{align}
, where $h$ stands for feature embedding space. Intuitively, in the fine-grained embedding space, each $v_{i,l}$ from $\mathcal{V}_i$ chooses the most relevant caption ${t_{i}^*}$ in the $\mathcal{T}_i$ using attention mechanism.  

Lastly, we project the fine-grained embedding space into the coarse-grained embedding space by modality fusion:
\begin{equation}
	h_{\mathcal{A}\mathcal{V}\mathcal{T}} = \mathcal{F}_{M}(h_{\mathcal{V}\mathcal{T}}, h_{\mathcal{A}})
\end{equation}
, where $\mathcal{F}_{M}$ denotes function of fusion and $h_{\mathcal{A}\mathcal{V}\mathcal{T}}$ represents coarse-grained feature representation. The complete process is summarized in Algorithm \ref{alg:mfst1}. An important note is though extracting the feature of three modalities has a cost of time, due to the transformer architecture and fusion operation, the cost of training and model size are not increased, compared to existing models using single or bimodal features.

\textbf{Multimodal Frame-Scoring Transformer.} Frame-scoring transformer \cite{NEURIPS2021_7503cfac} takes feature representation as input and predict importance scores. Note that previous approach exploits just text-visual representation, leaving audio features behind. MFST introduces frame-scoring transformer to video summarization, which is modified to predict frame-level importance scores $S$, based on coarse-grained feature representation $h_{\mathcal{A}\mathcal{V}\mathcal{T}}$:
\begin{align}
    \mathrm{Multimodal}~\text{-}~\mathrm{Attn.}(h_{\mathcal{A}\mathcal{V}\mathcal{T}}) = \mathrm{Concat}(\mathrm{h_1}, ..., \mathrm{h_h})W^{O_{\mathcal{A}\mathcal{V}\mathcal{T}}} \\
    \mathrm{h_i} = \mathrm{Attn.}(Q_{\mathcal{A}\mathcal{V}\mathcal{T}}W^{Q_{\mathcal{A}\mathcal{V}\mathcal{T}}}_i, K_{\mathcal{A}\mathcal{V}\mathcal{T}}W^{K_{\mathcal{A}\mathcal{V}\mathcal{T}}}_i, V_{\mathcal{A}\mathcal{V}\mathcal{T}}W^{V_{\mathcal{A}\mathcal{V}\mathcal{T}}}_i) \\
    \mathrm{Attn.}(h_{\mathcal{A}\mathcal{V}\mathcal{T}}) =\text{softmax}( \frac{Q_{\mathcal{A}\mathcal{V}\mathcal{T}}K_{\mathcal{A}\mathcal{V}\mathcal{T}}^{T}}{\sqrt{d_k}})V_{\mathcal{A}\mathcal{V}\mathcal{T}}
\end{align}
, where $W^{Q_{\mathcal{V}\mathcal{T}\mathcal{A}}}$, $W^{K_{\mathcal{V}\mathcal{T}\mathcal{A}}}$, and $W^{V_{\mathcal{V}\mathcal{T}\mathcal{A}}}$ denote parameter matrices and $d_k$ is the dimension of $K_{\mathcal{A}\mathcal{V}\mathcal{T}}$. 
 
Finally, we feed $h_{\mathcal{A}\mathcal{V}\mathcal{T}}$ to frame-scoring transformer with positional encodings at the bottom of transformer encoder and decoder stacks. Given ground truth frame scores $S_{gt}$ of N frames from a video, we train MFST using the mean square error:
\newcommand{\norm}[1]{\left\lVert#1\right\rVert}
\begin{equation}
\mathcal{L}_{\theta}(S_{gt}, S)=\frac{1}{N}\norm{S_{gt}-S}_2^2
\end{equation}
The complete process is summarized in Algorithm \ref{alg:mfst2}. 

\begin{algorithm}[t]
    \caption{Fine-to-Coarse Space Projection}
    \label{alg:mfst1}
    \begin{algorithmic}[1]
    \REQUIRE $V = \{V_1, V_2, ..., V_N\}$
    \FORALL{\begin{small}$V_i$\end{small}}
    \STATE Generate dense video captions $C_i = g_{c}(V_i)$ 
    \STATE , where $C_i = \{c^i_{1}, c^i_{2}, ..., c^i_{M}\}$
    \ENDFOR
    \FORALL{\begin{small}$V_i$ and $C_i$\end{small}}
    \STATE \begin{small}$\mathcal{T}_i = f_{t}(C_i)$, \end{small}
    \begin{small}$\mathcal{V}_i = f_{v}(V_i)$, and \end{small}
    \begin{small}$\mathcal{A}_i = f_{a}(V_i)$\end{small}
    \STATE , where $\mathcal{T} = \{\mathcal{T}_{1}, \mathcal{T}_{2}, ..., \mathcal{T}_{N}\}$, $\mathcal{V} = \{\mathcal{V}_{1}, \mathcal{V}_{2}, ..., \mathcal{V}_{N}\}$, and $\mathcal{A} = \{\mathcal{A}_{1}, \mathcal{A}_{2}, ..., \mathcal{A}_{N}\}$
    \ENDFOR
    \STATE \begin{small}$h_{\mathcal{V}\mathcal{T}}$ = $\text{Concat}(h_{\mathcal{V}_1\mathcal{T}_1}, h_{\mathcal{V}_2\mathcal{T}_2}$, ...,  $h_{\mathcal{V}_N\mathcal{T}_N)}$\end{small}
    \FORALL{\begin{small}$\mathcal{T}_i$, $\mathcal{V}_i$ and $\mathcal{A}_i$\end{small}}
    \STATE \begin{small}$h_{\mathcal{T}_i} := \mathcal{T}_i$,\end{small}
    \begin{small}$h_{\mathcal{V}_i} := \mathcal{V}_i$,\end{small}
    \begin{small}$h_{\mathcal{A}_i} := \mathcal{A}_i$,\end{small}
    \STATE Calculate fine-grained modality space \begin{small}$h_{\mathcal{V}_i\mathcal{T}_i}$\end{small}
    \STATE \begin{small}$h_{\mathcal{V}_i\mathcal{T}_i}$ = $\text{Attention}(h_{\mathcal{V}_i}, h_{\mathcal{T}_i})$\end{small}
    \STATE Project fine-to-coarse embedding space \begin{small}$h_{\mathcal{A}_i\mathcal{V}_i\mathcal{T}_i}$\end{small} 
    \STATE \begin{small}$h_{\mathcal{A}_i\mathcal{V}_i\mathcal{T}_i} = \mathcal{F}_{M}(h_{\mathcal{V}_i\mathcal{T}_i}, h_{\mathcal{A}_i})$\end{small} 
    \ENDFOR
    \STATE \begin{small}$h_{\mathcal{A}\mathcal{V}\mathcal{T}}$ = $\text{Concat}(h_{\mathcal{A}_1\mathcal{V}_1\mathcal{T}_1}, h_{\mathcal{A}_2\mathcal{V}_2\mathcal{T}_2}$, ...,  $h_{\mathcal{A}_N\mathcal{V}_N\mathcal{T}_N)}$\end{small}
    \end{algorithmic}
\end{algorithm}

\begin{algorithm}[t]
    \caption{Multimodal Frame-Scoring Transformer}
    \label{alg:mfst2}
    \begin{algorithmic}[1]
    \REQUIRE $h_{\mathcal{A}\mathcal{V}\mathcal{T}}$ = \{$h_{\mathcal{A}_1\mathcal{V}_1\mathcal{T}_1}, h_{\mathcal{A}_2\mathcal{V}_2\mathcal{T}_2}, ..., h_{\mathcal{A}_N\mathcal{V}_N\mathcal{T}_N}\}$
    \FOR{\begin{small}\text{each epoch}\end{small}}
    \FOR{\begin{small}$i$ in range(h)\end{small}}
    \STATE Calculate self-attention for each $i$
    \STATE \begin{small}$\text{Attn.}(h_{\mathcal{A}\mathcal{V}\mathcal{T}}) = \text{softmax}(Q_{\mathcal{A}\mathcal{V}\mathcal{T}}K_{\mathcal{A}\mathcal{V}\mathcal{T}}^{T} / \sqrt{d_k})V_{\mathcal{A}\mathcal{V}\mathcal{T}}$\end{small}
    \STATE Calculate single head attention $\text{h}_i$
    \STATE \begin{small}$\text{h}_i = \text{Attn.}(Q_{\mathcal{A}_i\mathcal{V}_i\mathcal{T}_i}, K_{\mathcal{A}_i\mathcal{V}_i\mathcal{T}_i}, V_{\mathcal{A}_i\mathcal{V}_i\mathcal{T}_i})$\end{small}
    \ENDFOR
    \STATE \begin{small}$\text{Multimodal-Attn.} = \text{Concat}(\text{h}_1, ..., \text{h}_h)$\end{small}
    \STATE \begin{small}$S = \text{Linear(Multimodal-Attn.)}$\end{small}
    \STATE Compute loss between ground truth scores $S_{gt}$ and predicted scores $S$
    \STATE \begin{small}$\mathcal{L}_{\theta}(S_{gt}, S)=\frac{1}{N}\norm{S_{gt}-S}_2^2$\end{small}
    \ENDFOR
    \end{algorithmic}
\end{algorithm}

\section{Experiments}

\begin{table}[t]
\definecolor{mygray-bg}{RGB}{211,211,211}
\caption{Experimental results on SumMe under the Canonical, Augment, and Transfer settings (F-score).}
\label{table:1}
\vspace{-3mm}
\center
\resizebox{\columnwidth}{!}{\begin{tabular}{l|ccc}
	\hline
	\multirow{2}{*}{Methods} & \multicolumn{3}{c}{SumMe} \\
	& \textit{Can} & \textit{Aug} & \textit{Tran} \\ \hline
	vsLSTM \cite{zhang2016video}& 0.376  & 0.416  & 0.407  \\
	SGAN \cite{mahasseni2017unsupervised}& 0.387  & 0.417 & --- \\
	SGAN$_{s}$ \cite{mahasseni2017unsupervised}& 0.417 & 0.436  &--- \\
	H-RNN \cite{zhao2017hierarchical}& 0.421 & 0.438  & --- \\
	DRDSN \cite{10.5555/3504035.3504964}& 0.421  & 0.439  & 0.426 \\
	HSA-RNN \cite{Zhao2018HSARNNHS} & 0.423 & 0.421  & --- \\
	ACGAN \cite{he2019unsupervised}&0.460 & 0.470  &0.445 \\
	WS-HRL \cite{chen2019weakly}&0.436 & 0.445  &--- \\
	re-S2S \cite{10.1007/978-3-030-01237-3_24} & 0.425 & 0.449  & --- \\
	S-FCN \cite{10.1007/978-3-030-01258-8_22}& 0.475 & 0.511 & 0.441 \\
	VASNet \cite{fajtl2018summarizing} & 0.497 & 0.510 & --- \\
	CSNet$_{s}$ \cite{DBLP:conf/aaai/JungCKWK19}&0.513 &  0.521 & 0.451 \\
	GLRPE \cite{10.1007/978-3-030-58595-2_11}&0.502 &  --- & --- \\
	SumGraph \cite{Park2020SumGraphVS} & 0.514 & 0.529 & 0.487 \\
	RSGN \cite{zhao2021reconstructive}&0.450 &0.457 &0.440\\
	RSGN$_{uns}$ \cite{zhao2021reconstructive}&0.423 &0.436 &0.412\\
	MSVA \cite{ghauri2021MSVA} & \underline{0.545}  & ---& --- \\
	CLIP-It \cite{NEURIPS2021_7503cfac} & 0.542 & \underline{0.564} & \underline{0.519} \\
	SSPVS \cite{https://doi.org/10.48550/arxiv.2201.02494} & 0.501 & ---& --- \\
	iPTNet \cite{jiang2022joint} & \underline{0.545}  & 0.569 & 0.492 \\
	 \cellcolor{mygray-bg}MFST (Ours) & \cellcolor{mygray-bg}\textbf{0.595}  &  \cellcolor{mygray-bg}\textbf{0.655} &  \cellcolor{mygray-bg}\textbf{0.576} \\
\hline
\end{tabular}}
\end{table}

\begin{table}[t]
\definecolor{mygray-bg}{RGB}{211,211,211}
\caption{Experimental results on TVSum under the Canonical, Augment, and Transfer settings (F-score).}
\label{table:3}
\vspace{-3mm}
\center
\resizebox{\columnwidth}{!}{\begin{tabular}{l|ccc}
	\hline
	\multirow{2}{*}{Methods} & \multicolumn{3}{c}{TVSum} \\
	& \textit{Can} & \textit{Aug} & \textit{Tran} \\ \hline
	vsLSTM \cite{zhang2016video}&0.542 & 0.579  & 0.569 \\
	SGAN \cite{mahasseni2017unsupervised}&  0.508 & 0.589 & ---  \\
	SGAN$_{s}$ \cite{mahasseni2017unsupervised}&0.563  &0.612  & --- \\
	H-RNN \cite{zhao2017hierarchical}& 0.579 & 0.619 &  --- \\
	DRDSN \cite{10.5555/3504035.3504964}& 0.581  & 0.598  & 0.589  \\
	HSA-RNN \cite{Zhao2018HSARNNHS}  & 0.587 & 0.598 &  ---  \\
	ACGAN \cite{he2019unsupervised}&0.585 & 0.589 & 0.578  \\
	WS-HRL \cite{chen2019weakly}& 0.584 &0.585  & ---  \\
	re-S2S \cite{10.1007/978-3-030-01237-3_24} & 0.603 & 0.639 & --- \\
	S-FCN \cite{10.1007/978-3-030-01258-8_22}& 0.568 &0.592 & 0.582 \\
	VASNet \cite{fajtl2018summarizing} & 0.614 & 0.623 & ---   \\
	CSNet$_{s}$ \cite{DBLP:conf/aaai/JungCKWK19} & 0.588 & 0.590 & 0.592  \\
	GLRPE \cite{10.1007/978-3-030-58595-2_11} & 0.591 & --- &  ---\\
	SumGraph \cite{Park2020SumGraphVS} & 0.639 & 0.658 &  0.605 \\
	RSGN \cite{zhao2021reconstructive}&0.601 &0.611 &0.600\\
	RSGN$_{uns}$ \cite{zhao2021reconstructive}&0.580 &0.591 &0.597\\
	MSVA \cite{ghauri2021MSVA} & 0.628& ---  & ---   \\
	CLIP-It \cite{NEURIPS2021_7503cfac} &  \underline{0.663} & \underline{0.690} & \underline{0.655}   \\
	SSPVS \cite{https://doi.org/10.48550/arxiv.2201.02494} &  0.607& ---  & ---   \\
	iPTNet \cite{jiang2022joint} & 0.634 & 0.642  & 0.598 \\
	 \cellcolor{mygray-bg}MFST (Ours) &  \cellcolor{mygray-bg}\textbf{0.737} &  \cellcolor{mygray-bg}\textbf{0.779} &  \cellcolor{mygray-bg}\textbf{0.691}   \\
\hline
\end{tabular}}
\end{table}

\begin{table}[t]
\definecolor{mygray-bg}{RGB}{211,211,211}
\caption{Experimental results on SumMe (Kendall’s $\tau$ and Spearman’s $\rho$). }\label{table:2}
\vspace{-3mm}
\center
\resizebox{\columnwidth}{!}{\begin{tabular}{l|cc}
\hline
\multirow{2}{*}{Methods} & \multicolumn{2}{c}{SumMe} \\
 & $\tau$ & $\rho$ \\ \hline
Random & ~0.000~ & ~0.000~  \\
Human & 0.205 & 0.213 \\
Ground Truth & 1.000 & 1.000  \\ \hline
SGAN \cite{mahasseni2017unsupervised}~ & --- & ---  \\
WS-HRL \cite{chen2019weakly} & --- & ---  \\
DRDSN \cite{10.5555/3504035.3504964} & 0.047 & 0.048  \\
dppLSTM \cite{zhang2016video} & --- & ---  \\
CSNet$_{s}$ \cite{DBLP:conf/aaai/JungCKWK19} & --- & ---  \\
GLRPE \cite{10.1007/978-3-030-58595-2_11} & --- & ---  \\
HSA-RNN \cite{Zhao2018HSARNNHS} & 0.064 & 0.066  \\
RSGN \cite{zhao2021reconstructive} & 0.083 & 0.085  \\ 
RSGN$_{u}$ \cite{zhao2021reconstructive} & 0.071 & 0.073  \\
SumGraph \cite{Park2020SumGraphVS} & --- & ---  \\
SSPVS \cite{https://doi.org/10.48550/arxiv.2201.02494} & 0.123 & 0.170 \\
MSVA \cite{ghauri2021MSVA} & \underline{0.200} & \textbf{0.230}  \\
\cellcolor{mygray-bg}MFST (Ours) & \cellcolor{mygray-bg}\textbf{0.229} & \cellcolor{mygray-bg}\underline{0.229} \\ \hline
\end{tabular}}
\end{table}

\begin{table}[t]
\definecolor{mygray-bg}{RGB}{211,211,211}
\caption{Experimental results on TVSum (Kendall’s $\tau$ and Spearman’s $\rho$). }\label{table:4}
\vspace{-3mm}
\center
\resizebox{\columnwidth}{!}{\begin{tabular}{l|cc}
\hline
\multirow{2}{*}{Methods} & \multicolumn{2}{c}{TVSum} \\
 & $\tau$ & $\rho$  \\ \hline
Random & ~0.000~ & ~0.000~  \\
Human  & 0.177 & 0.204 \\
Ground Truth & 0.364 & 0.456 \\ \hline
SGAN \cite{mahasseni2017unsupervised}~  & 0.024 & 0.032 \\
WS-HRL \cite{chen2019weakly}  & 0.078 & 0.116 \\
DRDSN \cite{10.5555/3504035.3504964}  & 0.020 & 0.026 \\
dppLSTM \cite{zhang2016video}  & 0.042 & 0.055 \\
CSNet$_{s}$ \cite{DBLP:conf/aaai/JungCKWK19}  & 0.025 & 0.034 \\
GLRPE \cite{10.1007/978-3-030-58595-2_11}  & 0.070 & 0.091 \\
HSA-RNN \cite{Zhao2018HSARNNHS}  & 0.082 & 0.088 \\
RSGN \cite{zhao2021reconstructive}  & 0.083 & 0.090 \\ 
RSGN$_{u}$ \cite{zhao2021reconstructive}  & 0.048 & 0.052 \\
SumGraph \cite{Park2020SumGraphVS} & 0.094 & 0.138 \\
SSPVS \cite{https://doi.org/10.48550/arxiv.2201.02494} & 0.169 & \textbf{0.231} \\
MSVA \cite{ghauri2021MSVA}  & \underline{0.190} & 0.210 \\
\cellcolor{mygray-bg}MFST (Ours) & \cellcolor{mygray-bg}\textbf{0.222} & \cellcolor{mygray-bg}\underline{0.224} \\ \hline
\end{tabular}}
\end{table}

\subsection{Two Fundamental Research Questions.} In our extensive experiments, we struggle with two fundamental research questions: \textit{(1) How to alleviate data sparsity problem?} and \textit{(2) How to learn diverse modality features to predict importance score better?} The first question arose because summarization datasets consist of frame-level human annotations and they are hard to collect. The second question should be solved in order to create a video summarization model considering multimodal information as people do. Note that these two questions are not separate, rather highly correlated. 

\subsection{Dataset Description}

We conduct our video summarization experiments on two benchmarks: \textbf{TVSum} \cite{7299154} and \textbf{SumMe} \cite{10.1007/978-3-319-10584-0_33}. 
\begin{itemize}
    \item TVSum \cite{7299154} contains 50 videos, including the topics of news, documentaries. The duration of each video varies from 1 to 10 minutes. 20 annotators provide frame-level importance scores for each video.  
    \item SumMe \cite{10.1007/978-3-319-10584-0_33} consists of 25 user videos, covering various topics (e.g., holidays and sports). Each video ranges from 1 to 6 minutes. 15 to 18 persons annotated multiple ground truth summaries for each video.
\end{itemize}

\subsection{Metric Description}
We follow the same experimental metrics used in existing works: F-score and Rank-based evaluation. \textit{True positive} means highlight overlaps between model-generated summary $V_{m}$ and human-generated summary $V_{h}$ based on importance scores. The precision and recall are calculated as follows: 
\begin{align}
    \text{Precision} = \frac{\mid V_m \cap V_h \mid}{\mid V_m \mid}, \text{Recall} = \frac{\mid V_m \cap V_h \mid}{\mid V_h \mid}.
\end{align}

Rank-based evaluations \cite{otani2019rethinking} compute Kendall's $\tau$ and Spearman's $\rho$ which measure non-parametric rank correlations: 
\begin{align}
    \tau &= Kendall(S_{gt}, S)\\
    \rho &= Spearman(S_{gt}, S)
\end{align}

Though Performance over Random (PoR) \cite{10.1145/3394171.3413632} proposed a new evaluation protocol for handling the non-overlapping splits, most existing methods did not disclose their code and experimented in the existing protocols which uses fixed number of test splits. Therefore, we follow the same experimental metrics and then evince the superiority of our model by a large margin.   

\subsection{Settings}
\textbf{Experimental Settings.} We compare MFST with existing models in three different experimental settings:
\begin{itemize}
    \item In Canonical setting, we selects the dataset (e.g., TVSum or SumMe) and randomly splits the dataset into training and evaluation.  
    \item In Augment setting, we merges the two datasets into one and randomly splits the dataset into training and evaluation. 
    \item In Transfer setting, we trains a model using one dataset and evaluates the trained model on the other dataset.
\end{itemize}
As we follow the experimental protocol in proposed by existing studies, in all experimental settings, we conduct experiments over 5 splits and average the results. Each experiment randomly selects 20\% of the dataset for evaluation.

\textbf{Implementation Details.} We leverage a Frame-Scoring Transformer with 8 heads, 6 encoder layers, and 6 decoder layers. We use a linear layer whose hidden dimension is 512.

Note that summarization datasets are hard to collect because they consist of frame-level human annotations. To mitigate data scarcity, we co-opt learned VideoCLIP \cite{xu-etal-2021-videoclip} to obtain both extracted features. We compute text-attended visual representation using attention mechanism, finding the most relevant caption per frame.
We leverage feature extractor using Wav2Vec2 \cite{NEURIPS2020_92d1e1eb} to exploit audio features. 

\textbf{Training Details.} We train our model on 8 NVIDIA GeForce TITAN GPUs for 20 epochs. The batch size are selected based on available GPU memory. We use Adam optimizer with learning rate 1e-4 and weight decay of 1e-3. Note that though extracting the feature of three modalities has a cost of time, due to the transformer architecture and fusion operation, the cost of training and model size are not increased, compared to existing models using single or bimodal features (not shown).   

\subsection{Performance Comparison}

\textbf{Results on Video Summarization.} We conduct experiments to answer the two questions. Table~\ref{table:1} and Table~\ref{table:2} show our extensive experiments with previous methods on SumMe dataset. Table~\ref{table:3} and Table~\ref{table:4} also demonstrates our comprehensive experiments with existing models on TVSum dataset. As shown in Table~\ref{table:1} and Table~\ref{table:3}, under the canonical, augment, and transfer settings, we demonstrate that MFST outperforms existing methods on the benchmark. Table~\ref{table:2} and Table~\ref{table:4} also show that MFST outperforms by a large margin in Rank-based evaluation except MSVA. However, MSVA did not conduct experiments in Augment setting and Transfer setting, it is hard to say MSVA is compatible with our model. 

Experimental results evince that methods which exploit two or more modality features and mitigate data sparsity problem get high a score than other methods. Among them, MFST, which properly exploits the visual, text and audio modalities, achieves state-of-the-art performance by a large margin in F1 score and Rank-based evaluation by nontrivial margins. Furthermore, thanks to the architecture of MFST, we speculate that MFST can circumvent \textit{data-hungry} and \textit{sensitive to the overfitting} issues, caused by the transformer structure. We leave the proof of this for future work.   

\subsection{Ablation Studies}

\textbf{Effect of Each Modality.} We further conduct ablation studies on SumMe and TVSum datsets. As shown in Table~\ref{table:5} and Table~\ref{table:6}, to validate the effect of our feature representation, we compare three different model variants, adding modality features. Results in Table~\ref{table:5} and Table~\ref{table:6} demonstrate our approach can be advanced when extracted modality features added. Interestingly, our framework without audio features also outperforms existing methods on both SumMe and TVSum. We conjecture that MFST exploiting pretrained model can represent modality features in a embedding space leading higher prediction performance than other models. Though existing methods get higher score as they use more modality information than counterparts, most of them did not disclose the code, so we leave ablations about modality using them for future work. In the following section, we show qualitative analysis and ablation.   

\begin{table}[H]
\definecolor{mygray-bg}{RGB}{255,218,185}
\caption{Effect of each modality feature on SumMe (F-score).}
\label{table:5}
\vspace{-3mm}
\center
{\begin{tabular}{ccc|ccc}
	\hline
	\multicolumn{3}{c|}{Modality} & \multicolumn{3}{c}{SumMe} \\ 
	$\mathcal{V}$ & $\mathcal{T}$ & $\mathcal{A}$ & \textit{Tran} & \textit{Can} & \textit{Aug}\\ \hline
	\textcolor[RGB]{0,128,0}{\ding{51}}  & \textcolor[RGB]{178,34,34}{\ding{55}}  & \textcolor[RGB]{178,34,34}{\ding{55}} & 0.525 & 0.564  & 0.505 \\
	\textcolor[RGB]{0,128,0}{\ding{51}}  & \textcolor[RGB]{0,128,0}{\ding{51}}  & \textcolor[RGB]{178,34,34}{\ding{55}} & 0.542 & 0.629  & 0.553 \\ \hline
    \textcolor[RGB]{0,128,0}{\ding{51}}  & \textcolor[RGB]{0,128,0}{\ding{51}}  & \textcolor[RGB]{0,128,0}{\ding{51}} & 0.595 & 0.655  & 0.576 \\ 
\hline
\end{tabular}}
\end{table}

\begin{table}[H]
\definecolor{mygray-bg}{RGB}{255,218,185}
\caption{Effect of each modality feature on TVSum (F-score).}
\label{table:6}
\vspace{-3mm}
\center
{\begin{tabular}{ccc|ccc}
	\hline
	\multicolumn{3}{c|}{Modality} & \multicolumn{3}{c}{TVSum} \\ 
	$\mathcal{V}$ & $\mathcal{T}$ & $\mathcal{A}$ & \textit{Can} & \textit{Aug} & \textit{Tran} \\ \hline
	\textcolor[RGB]{0,128,0}{\ding{51}}  & \textcolor[RGB]{178,34,34}{\ding{55}}  & \textcolor[RGB]{178,34,34}{\ding{55}} & 0.659 & 0.674  & 0.629 \\
	\textcolor[RGB]{0,128,0}{\ding{51}}  & \textcolor[RGB]{0,128,0}{\ding{51}}  & \textcolor[RGB]{178,34,34}{\ding{55}} & 0.708 & 0.753  & 0.659 \\ \hline
    \textcolor[RGB]{0,128,0}{\ding{51}}  & \textcolor[RGB]{0,128,0}{\ding{51}}  & \textcolor[RGB]{0,128,0}{\ding{51}} & 0.737 & 0.779  & 0.691 \\
\hline
\end{tabular}}
\end{table}

\begin{figure*}[t]
	\centering
	\begin{minipage}[h]{\linewidth}
		\centering
		\includegraphics[width=1\linewidth]{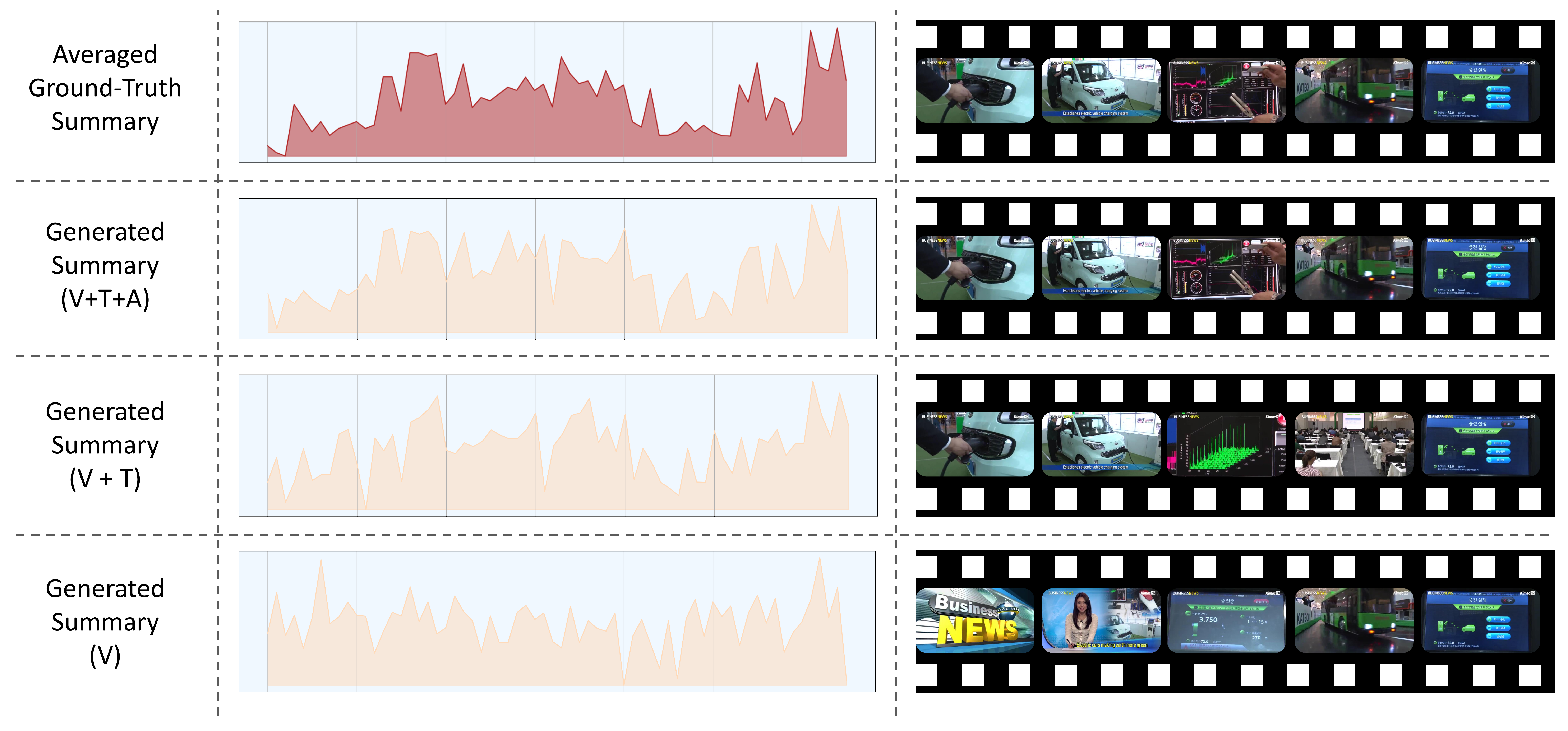}
	\end{minipage}
	\figvspace
	\caption{Comparison of \textcolor[RGB]{255,218,185}{predicted scores} and averaged \textcolor[RGB]{178,34,34}{ground truth scores} from TVSum dataset. Note that MFST-predicted score graph and ground truth graph are fairly similar in that maxima and minima.}
	\vspace{-3mm}
	\label{fig:qual}
\end{figure*}

\subsection{Qualitative Results.} In Figure~\ref{fig:qual}, we visualize importance scores generated by MFST and ground-truth scores (left), and video summarization results (right). While comparing the summary results, we add each modality information sequentially to our frame-scoring transformer when predicting the importance score. Interestingly, we observe that MFST generates a summary considering the modality information whenever each additional modality features are used. For example, the summary whose model uses three modalities focuses on frames to check electric voltage with \textit{sound of iron hitting} while the other summaries whose model uses one or two modalities fail to capture the highlight frames. As shown in Figure~\ref{fig:qual}, MFST predicts importance score similar to the ground truth score as more modality information is exploited. Indeed, model-generated score graph and ground truth graph are very similar, which means maximum points and minimum points in both graphs are fairly overlapped. Results in Figure~\ref{fig:qual} represents our model predicts parts that humans find interesting or not.

\section{Conclusion}

The conclusion of this paper is threefold:

\begin{itemize}
    \item We propose MFST, a simple and effective frame-scoring framework given videos. MFST exploits audio-visual-text features using learned feature extractors and frame-scoring multimodal transformer. MFST first generates caption-attended representation on fine-grained embedding space using attention mechanism. Then, our model project fine-grained space to coarse-grained space based on modality fusion.  
    \item We wrestle with two underlying research questions for generic video summarization: \textit{(1) How to alleviate data sparsity problem?} and \textit{(2) How to learn diverse modality features to predict importance score better?} The first question arose because summarization datasets consist of frame-level human annotations and they are hard to collect. The second question should be solved in order to create a video summarization model considering multimodal information as people do.
    \item Our comprehensive comparisons with previous approaches and ablation studies on generic video summarization datasets (TVSum and SumMe) evince that MFST can surpass its all counterparts by nontrivial margins, and attests the effectiveness and superiority of our approach.
\end{itemize}

\section{Discussion and Future Work}

\textbf{Limitations.} Though we demonstrate the effectiveness and superiority of our proposed method, lack of deep insights about modality representations and utilizing large-scale models are biggest limitations of our work. For example, we could not discover precise reasons why our framework is superior with respect to modality representations.

Additionally, This work also lacks the scrutiny to represent the modality well in the common space. In other words, we recognize that this work doesn't provide experiments using various model architectures to reflect modality information to our model.

Despite these limitations, it is clear that our approach handle three different modalities well and achieves state-of-the-art performance by a large margin in F1 score and Rank-based evaluation by nontrivial margins.   

\textbf{Future Works.} Our direction of future work is threefold:

\begin{itemize}
    \item Proving that our framework can circumvent data-hungry and sensitive to the overfitting issues, caused by the transformer structure, both theoretically and empirically.  
    \item Further study to express each modality information well in common space (e.g., \textit{contrastive learning}).
    \item Further research and ablations using existing approaches and various model architectures to better reflect modality information.  
\end{itemize}






\bibliographystyle{named}
\bibliography{ijcai23}

\end{document}